\documentclass[pdflatex,sn-mathphys-num]{sn-jnl}

\usepackage[utf8]{inputenc} 
\usepackage[T1]{fontenc}    
\usepackage{graphicx}       
\usepackage{subcaption}     
\usepackage{multirow}       
\usepackage{amsmath,amssymb,amsfonts} 
\usepackage{amsthm}         
\usepackage{mathrsfs}       
\usepackage{xcolor}         
\usepackage{textcomp}       
\usepackage{booktabs}       
\usepackage{appendix}       
\usepackage{fancyhdr}       
\usepackage{algorithm}      
\usepackage{algorithmicx}
\usepackage{algpseudocode}  
\usepackage{array}
\usepackage{makecell}
\usepackage{listings}       
\usepackage{nicefrac}       
\usepackage{microtype}      
\usepackage{url}            
\usepackage{hyperref}       
\usepackage{manyfoot}

\theoremstyle{thmstyleone}%
\theoremstyle{thmstyletwo}%

\theoremstyle{thmstylethree}%

\begin{document}

\title{\textbf{ROAD}: \textbf{R}esponsibility-\textbf{O}riented Reward Design for Reinforcement Learning in \textbf{A}utonomous \textbf{D}riving}


\author[1]{\fnm{Yongming} \sur{Chen}}

\author[2]{\fnm{Miner} \sur{Chen}}

\author[1]{\fnm{Liewen} \sur{Liao}}

\author[3]{\fnm{Mingyang} \sur{Jiang}}

\author[4]{\fnm{Xiang} \sur{Zuo}}

\author[1]{\fnm{Hengrui} \sur{Zhang}}

\author[1]{\fnm{Yuchen} \sur{Xi}}

\author*[1]{\fnm{Songan} \sur{Zhang}}\email{songanz@sjtu.edu.cn}

\affil*[1]{\orgdiv{Global Institute of Future Technology}, \orgname{Shanghai Jiao Tong University}, \orgaddress{\street{800 Dongchuan Road}, \city{Shanghai}, \postcode{200240}, \country{China}}}

\affil[2]{\orgdiv{College of Humanities}, \orgname{Donghua University}, \orgaddress{\street{1882 West Yan'an Road}, \city{Shanghai}, \postcode{200051}, \country{China}}}

\affil[3]{\orgdiv{Department of Automation}, \orgname{Shanghai Jiao Tong University}, \orgaddress{\street{800 Dongchuan Road}, \city{Shanghai}, \postcode{200240}, \country{China}}}

\affil[4]{\orgdiv{School of Aeronautics and Astronautics}, 
\orgname{Shanghai Jiao Tong University}, \orgaddress{\street{800 Dongchuan Road}, 
\city{Shanghai}, 
\postcode{200240}, 
\country{China}}}


\abstract{Reinforcement learning (RL) in autonomous driving employs a trial-and-error mechanism, enhancing robustness in unpredictable environments. However, crafting effective reward functions remains challenging, as conventional approaches rely heavily on manual design and demonstrate limited efficacy in complex scenarios. To address this issue, this study introduces a responsibility-oriented reward function that explicitly incorporates traffic regulations into the RL framework. Specifically, we introduced a Traffic Regulation Knowledge Graph and leveraged Vision-Language Models alongside Retrieval-Augmented Generation techniques to automate reward assignment. This integration guides agents to adhere strictly to traffic laws, thus minimizing rule violations and optimizing decision-making performance in diverse driving conditions. Experimental validations demonstrate that the proposed methodology significantly improves the accuracy of assigning accident responsibilities and effectively reduces the agent's liability in traffic incidents.}




\maketitle

\section{Introduction}\label{sec1}
Among the learning-based methods for autonomous driving decision-making, imitation learning and reinforcement learning are two of the most widely studied paradigms. Although imitation learning effectively learns human driving behaviors and performs reliably in common scenarios, it often struggles with rare or long-tailed events due to the limited coverage of complex conditions in the training data \cite{huang2024intermediate,10554639,competitivebehavior2024qian}. In contrast, reinforcement learning leverages interaction with the environment to continuously explore and adapt, allowing better generalization to unseen situations \cite{shan2020reinforcement,xu2024multi,chen2024exploration}. This inherent capacity to adapt through continuous interaction enables RL-based autonomous driving systems to achieve superior robustness and decision-making performance in long-tailed scenarios, thus significantly improving safety, flexibility, and responsiveness in complex and dynamic road environments.

However, a central challenge in RL-based autonomous driving lies in the design of reward functions that can effectively guide agent behavior in diverse and complex road scenarios \cite{decisionmaking2024wang}. Hand-made rewards demand labor intensive tuning and still struggle to cover the myriad edge cases that arise on real roads \cite{knox2022rewardmisdesignautonomousdriving}. To mitigate these limitations, recent efforts have turned to Vision-Language Models (VLMs) to dynamically infer contextual rewards \cite{venuto2024code,wang2024rl,huang2024vlm}. Although VLMs provide greater generalization through multi-modal understanding, they are also prone to hallucinations \cite{zhu2024ibd,huo2024selfintrospectivedecodingalleviatinghallucinations,liu2024survey}.

To address these hallucination risks and enhance the reliability of reward inference, one promising direction is the integration of traffic regulations into the reward function. Within the RL framework for autonomous driving, compliance with traffic regulations is critical to legal operation on public roads \cite{MARTINHO2021556,Gill2021,doi:10.2214/AJR.21.27224}. Integrating traffic rules into the design of the reward function ensures that agents inherently adhere to legal and ethical standards during decision making. Regulations such as giving priority to oncoming traffic when making a left turn and giving priority to vehicles circulating within roundabouts reflect a social consensus that places safety above convenience. Proper enforcement of these regulations significantly mitigates accident risks, maintains traffic order, and aligns decision-making processes with human-centric values, improving societal acceptance. Thus, a reward function designed to enforce compliance with traffic laws is not only technically necessary but also ethically and legally imperative, fostering trustworthy autonomous systems and public confidence.

\begin{figure}[h]
  \centering
  \includegraphics[width=1\textwidth]{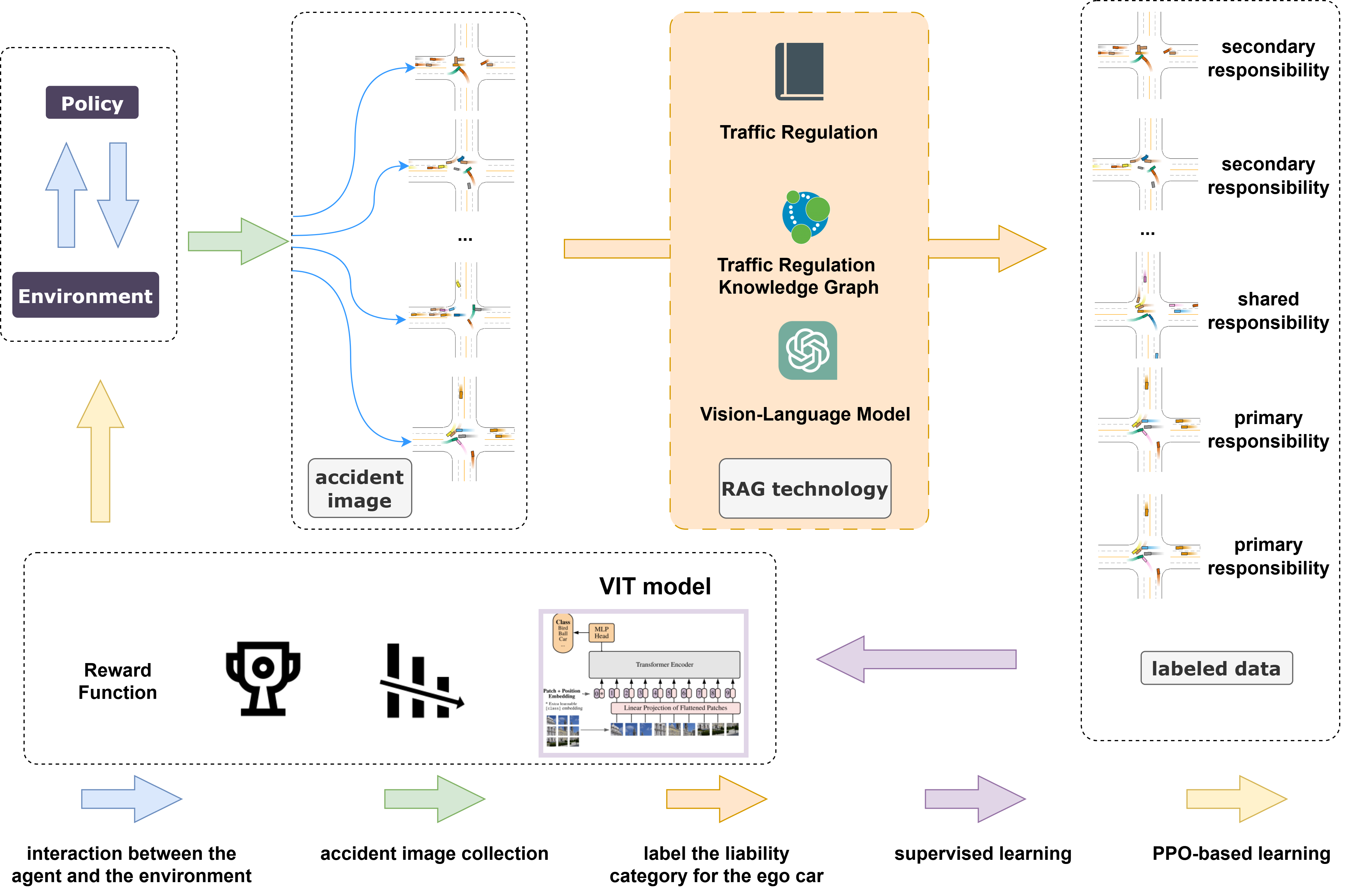}  
  \caption{Overview of the responsibility-aware reinforcement learning pipeline. Simulated accidents are generated to produce collision images, which are labeled with responsibility levels using VLM and RAG techniques. The labeled dataset trains a responsibility classifier, a VIT\cite{dosovitskiy2020image} model, whose output informs the reward function for policy learning.}
  \label{fig:Fig1}
\end{figure}

This paper proposes \textbf{ROAD} (\textbf{R}esponsibility-\textbf{O}riented Reward Design for Reinforcement Learning in \textbf{A}utonomous \textbf{D}riving ), a method that incorporates traffic regulations into the reward function to promote responsible behavior in reinforcement learning for autonomous driving, as shown in Fig. \ref{fig:Fig1}. Our key insight is that assigning rewards based on the degree of responsibility in traffic accidents can effectively align autonomous agents' decision-making incentives with traffic regulation standards. Specifically, we quantify driving appropriateness through post-collision responsibility, assigning greater penalties to actions associated with higher liability. To ensure accurate and consistent responsibility assessment while mitigating hallucinations from Vision Language Models (VLM), we constructed a structured Traffic Regulation Knowledge Graph (TRKG) distilled from authoritative legal documents. By integrating TRKG with VLM and Retrieval-Augmented Generation (RAG), we produce a diverse set of labeled traffic incident cases. These are used to train an offline responsibility classification model, avoiding reliance on real-time API calls.  Ultimately, the responsibility predictions from this offline model inform the reward mechanism, significantly enhancing both compliance and generalization capabilities of trained autonomous driving agents.\\

In summary, our contributions include the following.
\begin{enumerate}
\item Addressing hallucination problems in VLM by designing a specialized TRKG ontology for accident liability determination, along with an automated knowledge graph construction pipeline that improves response accuracy.
\item Enhancing the consistency of reward function design through an accident responsibility assignment model that integrates TRKG and VLM, combined with an automated reward generation approach.
\item We used the Road Traffic Safety Law of the People's Republic of China \cite{gov} to demonstrate that autonomous driving models trained in the MetaDrive simulator \cite{li2022metadrive} achieve higher success rates and a lower degree of accident liability.
\end{enumerate}

\section{RELATED WORK}
\label{sec:rw}

\subsection{Applications of Visual-Language Models in Autonomous Driving}

Recent advances in the application of visual language models (VLMs) to autonomous driving have demonstrated substantial potential. By effectively integrating visual and linguistic modalities, VLMs enhance understanding of intricate road scenarios and traffic regulations, thus increasing the intelligence and reliability of autonomous driving decisions. Marcu et al. \cite{10.1007/978-3-031-72980-5_15} introduced the LingoQA model, explicitly designed for visual question-answering tasks within autonomous driving contexts. This model leverages multi-modal input to enhance decision-making capabilities in complex traffic environments. Additionally, VLMs exhibit notable efficacy in managing long-tail data challenges. Fu et al. \cite{fu2023drivelikehumanrethinking} demonstrated that VLMs substantially contribute to achieving human-like driving behaviors, particularly in rare scenarios inadequately captured by conventional datasets, by effectively learning from limited data samples.

Despite recent advances, the use of VLMs for reward function design in autonomous driving remains largely unexplored, with existing research primarily targeting perception and planning rather than explicit modeling of reward function.

\subsection{Comparison Between Imitation Learning and Reinforcement Learning in Autonomous Driving}

Two mainstream learning paradigms commonly used in autonomous driving tasks are Imitation Learning (IL) and Reinforcement Learning (RL). IL learns policies by mimicking expert demonstrations, with the objective of minimizing the discrepancy between the model's actions and those of the expert. In contrast, RL continuously interacts with the environment to collect feedback and optimizes the policy based on a pre-defined reward function.

Although IL can efficiently learn human-like driving strategies from large-scale expert data, it suffers from notable limitations in terms of generalization and fault tolerance. Specifically, IL policies are highly dependent on the distribution of training data and often suffer substantial performance drops in novel environments due to distributional shifts between training and test scenarios \cite{huang2024intermediate, 10554639, competitivebehavior2024qian}. Furthermore, traditional IL models typically lack mechanisms to estimate the uncertainty of the decision and cannot provide confidence levels for predicted behaviors, which makes it difficult to adjust or correct the policy when encountering ambiguous or unknown traffic scenarios \cite{delavari2025caril}. These limitations lead to poor performance of pure IL-based autonomous driving systems when dealing with out-of-distribution edge cases, uncertain environments, or tasks without expert demonstrations \cite{ye2024lordlargemodelsbased}.

In contrast, RL demonstrates multiple advantages in autonomous driving scenarios. RL has strong autonomous exploration capabilities. By performing trial-and-error in simulated environments, agents can independently discover effective driving strategies that adapt to complex and dynamically changing traffic conditions \cite{shan2020reinforcement}. In addition, RL allows for flexible reward design. Researchers can customize the reward function according to task requirements, incorporating multiple objectives such as driving safety, ride comfort, and travel efficiency to achieve balanced decision making \cite{xu2024multi, chen2024exploration}. This tunable reward mechanism enables agents to learn strategies aligned with practical driving demands. Moreover, RL optimizes policies based on clearly defined reward signals and performs closed-loop training during interaction with the environment, which contributes to stronger robustness of the learned strategies against environmental disturbances and distributional shifts \cite{lu2023imitation}.

In summary, reinforcement learning effectively addresses the shortcomings of imitation learning in terms of generalization and safety performance. It provides explicit objectives and feedback mechanisms for policy design and optimization, enabling autonomous agents to acquire safer and more reliable driving behaviors.

\subsection{Reinforcement Learning with Visual-Language Model (VLM) Feedback}
Integrating VLM-generated feedback into reinforcement learning can significantly enhance a model's capacity to interpret complex scenarios due to the simultaneous processing of visual and linguistic data, resulting in richer and more precise feedback. Venuto et al. \cite{venuto2024code} leveraged VLMs to divide complex tasks into subtasks, generating validation code to translate sparse feedback into dense and actionable insights, thus improving the efficiency and effectiveness of RL agents. Wang et al. \cite{wang2024rl} used VLMs to assess task performance in robotic arms, utilizing this feedback to train reward models capable of accurately evaluating success. Similarly, Huang et al. \cite{huang2024vlm} proposed the "Contrasting Language Goal as Reward" framework, in which VLMs calculate similarity metrics between input imagery and positive or negative language goals to form comprehensive reward signals. These studies collectively demonstrate that VLMs serve as a powerful and versatile tool for bridging the gap between perception and decision-making, substantially enhancing reasoning capacity and reward modeling quality in reinforcement learning.

However, a notable limitation of VLMs is their susceptibility to "hallucinations," where minor inaccuracies in the outputs can substantially degrade model performance \cite{zhu2024ibd,huo2024selfintrospectivedecodingalleviatinghallucinations,liu2024survey}. This issue critically affects reinforcement learning applications, as feedback reliability directly influences policy quality. To mitigate this, incorporating structured external knowledge, such as knowledge graphs, via retrieval-augmented generation (RAG) has been shown to effectively guide the VLM's outputs, improving factual grounding and reducing the prevalence of hallucinations \cite{Hu2023CVPR}.

\section{OUR METHOD}
To mitigate the hallucination problem of VLMs and to guide autonomous vehicles in complying with traffic regulations during navigation, we construct a structured Traffic Regulation Knowledge Graph (TRKG) in Section \ref{sec31} to systematically store and represent relevant legal information. In Section \ref{sec32}, we train an automated traffic accident responsibility assignment model by integrating TRKG with VLM. Section \ref{sec34} then details the design of a responsibility-aware reward function based on the assigned responsibilities, thereby embedding regulatory compliance into the reinforcement learning training process.

\subsection{LLM-Based TRKG Construction}
\label{sec31}
\begin{figure}[h]
  \centering
  \includegraphics[width=0.9\textwidth]{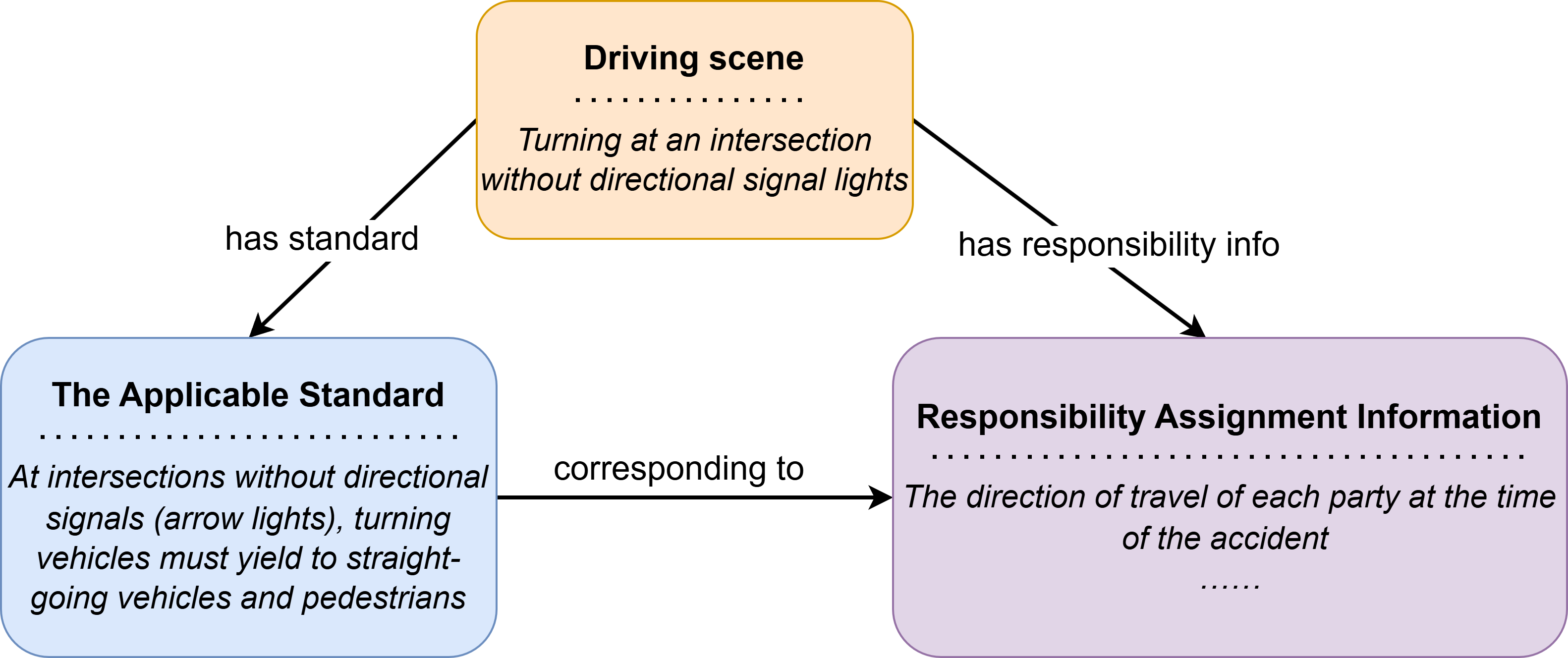}  
  \caption{The ontology structure of TRKG. Each box represents a node, with the upper part indicating the node type and the lower part presenting an example.}
  \label{fig:Fig2}
\end{figure}

As shown in Fig. \ref{fig:Fig2}, the ontology structure of the Traffic Regulation Knowledge Graph(TRKG) consists of three types of nodes: Driving Scene, The Applicable Standard, and Responsibility Assignment Information. The Driving Scene node stores various driving scenarios, such as intersections and roundabouts. The Applicable Standard node contains several rules for assigning responsibility, which are derived from traffic regulations. The Accident Scene Information node stores essential accident scene details to complete the responsibility assignment task. For example, in a cross-road scenario, one of the Applicable Standards is: “A vehicle making a left turn must yield to oncoming vehicles in the opposite lane'. To determine the responsibility in this scenario, the following two pieces of information are required: "The direction of travel of each party at the time of the accident" and "The speed of each party at the time of the accident".

\begin{table}[ht]
\caption{Prompts for Extracting Nodes in Traffic Regulation Knowledge Graph}
\label{table1}
\begin{tabular}{m{2cm}|m{4cm}|m{5cm}}
\textbf{Node Type} & 
\textbf{Prompt} & 
\textbf{Example Answer} \\
\hline
Driving Scene & 
What driving scenarios does this law apply to? & 
Turning at an intersection without directional signal lights \\
\hline
The Applicable Standard & 
What are the criteria for determining responsibility in this scenario under this law? & 
Turning vehicles must yield to straight-going vehicles and pedestrians when no directional signals (arrow lights) are present \\
\hline
Responsibility Assignment Information & 
What traffic accident scene information is needed to determine responsibility under this law? & 
The direction of travel of each party at the time of the accident \\
\end{tabular}
\end{table}

The manual construction of TRKG for responsibility assignment is both time-intensive and laborious. To address this limitation, we employ LLM through a prompt-based automation framework, as illustrated in Table \ref{table1}. For each regulation, the prompts are designed to extract three key node types: Driving Scene, Applicable Standard, and Accident Scene Information Required for Responsibility Assignment, thus enabling the systematic and efficient construction of a comprehensive TRKG.

\subsection{Automated Accident Responsibility Assignment Model Based on TRKG and VLM}
\label{sec32}

\begin{figure}[h]
  \centering
  \includegraphics[width=0.7\textwidth]{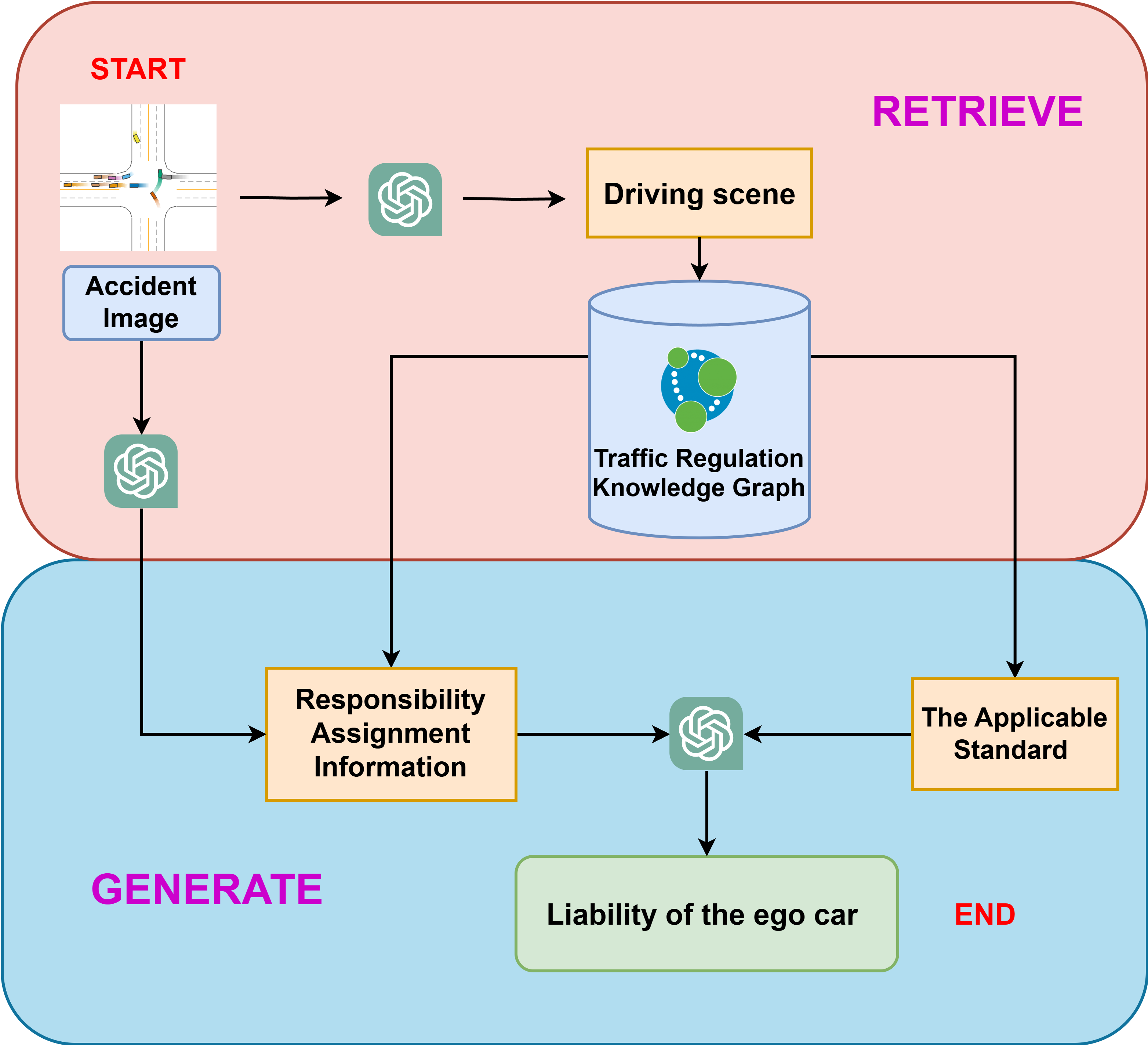}  
  \caption{Automated Accident Responsibility Assignment Model}
  \label{fig:Fig3}
\end{figure}

As illustrated in Fig. \ref{fig:Fig3}, this study proposes an automated responsibility assignment framework based on the chain-of-thought reasoning paradigm \cite{wei2023chainofthoughtpromptingelicitsreasoning}. The process comprises four main stages: identifying the driving scene using VLM, retrieving relevant standards and scene details from a traffic regulation knowledge graph, extracting accident details directly from images via VLM, and finally inferring responsibility assignments using an LLM. Decoupling visual information extraction from responsibility inference reduces potential inaccuracies from hallucinated scene elements. Table \ref{table2} details the prompts used for assignment tasks of responsibility. 

\begin{table}[ht]
\caption{Tasks and Prompts for Accident Responsibility Assignment}
\label{table2}
\centering
\begin{tabular}{m{3.5cm}|m{8cm}}
\textbf{Task} & \textbf{Prompt} \\
\hline
Scene identification & Please analyze the image provided and describe the scene of the traffic accident. Identify the location (e.g., intersection, highway), and any visible environmental factors (e.g., weather, road conditions). Provide a brief overview of the accident's context and any relevant details about the scene. \\
\hline
Scene info extraction & Please answer the following questions one by one according to the image provided: \{“The contents of the accident scene information required for assigning responsibility obtained in step 2”\} \\
\hline
Responsibility assignment & Here are the following traffic rules and the following factual information, please determine how much responsibility the ego car is responsible for. (On a scale of 0-10, where 0 means no responsibility and 10 means full responsibility.) \\
\end{tabular}
\end{table}

To address the high computational costs associated with frequent calls to OpenAI’s VLM \cite{openai_gpt4o_2024}, this study introduces an offline neural network model as a cost-effective alternative. Collision images depicting interactions between the ego vehicle and other agents were collected from simulation scenarios and annotated according to responsibility levels (primary, secondary, or shared). This annotated data set was later used to train a responsibility classification neural network employing supervised learning. Data augmentation techniques were applied to the training set to improve model generalization and classification performance.

\subsection{Reinforcement Learning Reward Function Design Based on Automated Accident Responsibility Assignment Model}
\label{sec34}

\begin{figure}[h]
  \centering
  \includegraphics[width=0.6\textwidth]{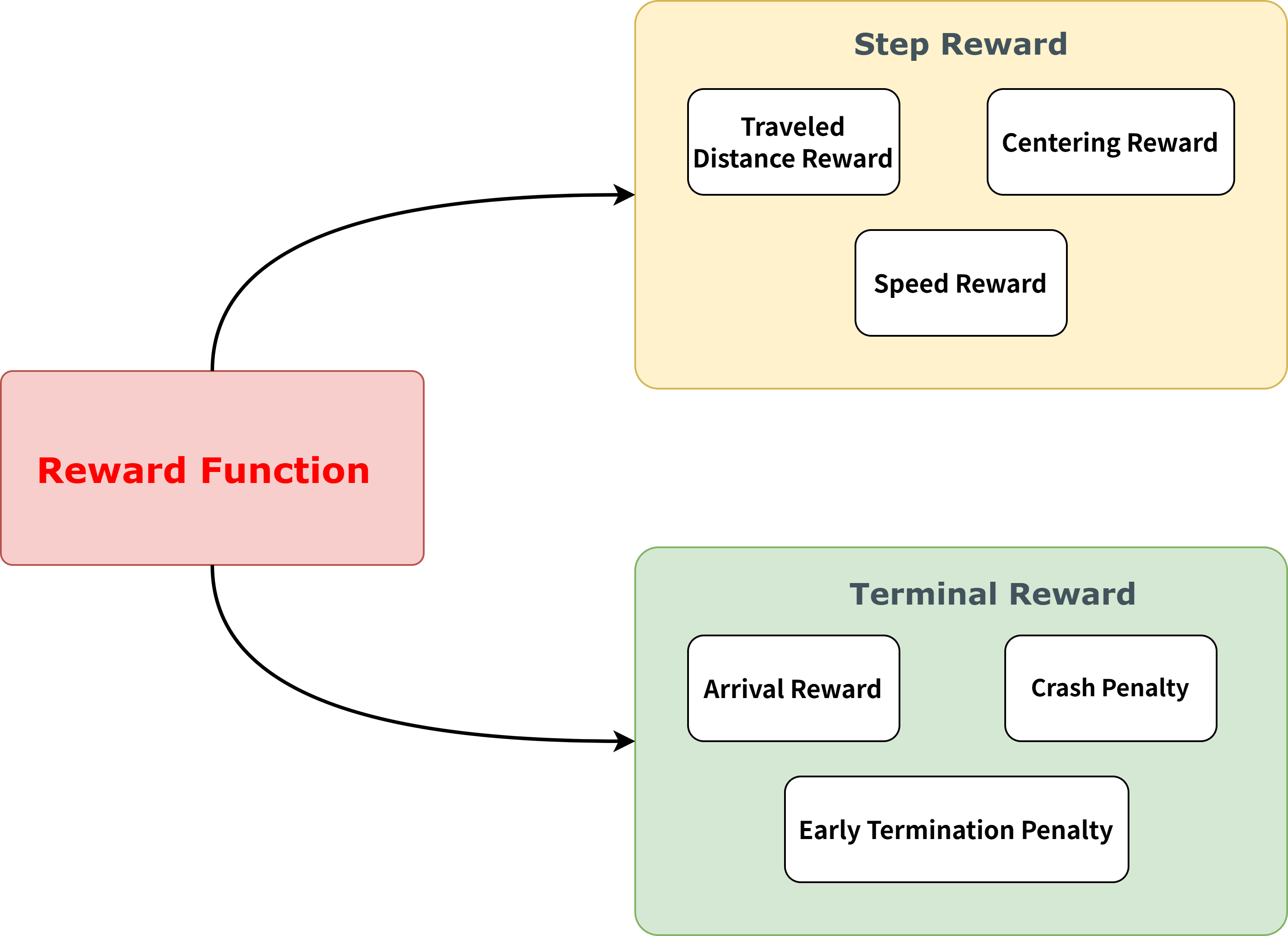}  
  \caption{Structure of Reinforcement Learning Reward Function}
  \label{fig:Fig5}
\end{figure}

As shown in Fig. \ref{fig:Fig5}, the reward function of our model is composed of a step reward and a terminal reward. The step reward includes the following components: Forward Reward, Centering Reward, and Speed Reward. The terminal reward includes the following components: Arrival Reward, Crash Penalty, and Early Termination Penalty. Among these, only the Crash Penalty is modified in our approach, while the other components remain unchanged from the original MetaDrive \cite{li2022metadrive} reward function.
\\

The \textbf{Crash Penalty} in the baseline model is fixed ($P_{\text{baseline}}$), while in our method it is computed dynamically:
\begin{equation}
r_{\text{crash}} = \lambda \times P_{\text{baseline}} \times \text{Accident\_liability\_ratio}
\end{equation}

where $\lambda$ is a scaling factor ensuring consistency with the baseline and Accident\_liability\_ratio is the responsibility that the ego car should bear in an accident.

The \textbf{Traveled Distance Reward} incentivizes vehicle progression by rewarding the distance traveled between consecutive states:
\begin{equation}
r_{\text{distance}} = \alpha_d \times \text{distance\_forward}
\end{equation}
where $\alpha_d$ is the distance reward scaling factor.

The \textbf{Centering Reward} encourages lane adherence by assigning rewards based on how closely the vehicle remains centered within the lane:
\begin{equation}
r_{\text{centering}} = \alpha_c \times \left( 1 - \frac{\text{deviation\_dist}}{\text{road\_width}} \right)
\end{equation}
where $\alpha_c$ is the centering reward coefficient and deviation\_dist is the dist from the road center.

Additionally, the \textbf{Speed Reward} promotes adherence to speed limits:
\begin{equation}
r_{\text{speed}} = \alpha_s \times \frac{\text{speed}}{\text{speed\_max}}
\end{equation}
where $\alpha_s$ is the speed reward coefficient, speed is the speed of ego car and speed\_max is the maximum speed limit of the road.

To reinforce successful task completion, an \textbf{Arrival Reward} $R_{\text{arrival}}$ is given.

Finally, a fixed \textbf{Early Termination Penalty} $P_{\text{early termination}}$ is added to the reward signal whenever an episode terminates early.

\section{Experiment}
\subsection{Experiment setup}
This study utilizes MetaDrive \cite{li2022metadrive} for simulation training, specifically employing two scenarios, which combine intersections (X) and roundabouts (O) to thoroughly test driving policies under diverse traffic conditions. Reinforcement learning (RL) is performed using the Proximal Policy Optimization (PPO) algorithm \cite{schulman2017proximal} provided by the Stable-Baselines3 \cite{schulman2017proximal} library. The hyperparameters of the environment, the PPO algorithm, and the reward function during the training process are detailed in Table \ref{table3}. In the intersection scenario, the total steps of the hyperparameter is set to $1.8 \times 10^{6}$, while in the roundabout scenario, it is increased to $2.4 \times 10^{6}$ to ensure model convergence. For Accident\_liability\_ratio, we set 1 for primary responsibility, 0.5 for shared responsibility, and 0 for secondary responsibility.

\begin{table}[ht]
\centering
\caption{Hyperparameters for the Environment, PPO Algorithm, and Reward Function}
\label{table3}
\begin{tabular}{p{3.8cm}|l|l}
\textbf{Category} & \textbf{Name} & \textbf{Value} \\ \hline
\multirow{6}{*}{\makecell[l]{\textbf{Environmental}\\\textbf{Hyperparameters}}}
    & random seed                  & 0                     \\ 
    & map                          & X\symbol{92}O         \\ 
    & random spawn lane index     & TRUE                  \\ 
    & num scenarios                & 2000                  \\ 
    & start seed                  & 5                     \\ 
    & traffic density              & 0.5                   \\ \hline
\multirow{6}{*}{\makecell[l]{\textbf{PPO Algorithm}\\\textbf{Hyperparameters}}}
    & n steps                      & 4096                  \\ 
    & verbose                      & 1                     \\ 
    & device                       & CPU                   \\ 
    & CheckpointCallback           & 10000                 \\ 
    & learning rate                & 0.0003                \\ 
    & total step                   & \texttt{1.8M\symbol{92}2.4M} \\ \hline
\multirow{10}{*}{\makecell[l]{\textbf{Reward Function}\\\textbf{Hyperparameters}}}
    & $\alpha_d$ (distance reward coeff.)       & 0.1    \\
    & $\alpha_c$ (centering reward coeff.)      & 0.01   \\
    & $\alpha_s$ (speed reward coeff.)          & 0.1    \\
    & $R_{\text{arrival}}$ (arrival reward)     & 40     \\
    & $P_{\text{baseline}}$ (baseline crash penalty) & -20 \\
    & $\lambda$ (crash penalty scaling factor)   & 1 \\
    & $P_{\text{early termination}}$ (early termination penalty) & -20 \\
\end{tabular}
\end{table}

\subsection{Traffic Regulations Knowledge Graph Construction}
\begin{figure}[h]
  \centering
  \includegraphics[width=0.8\textwidth]{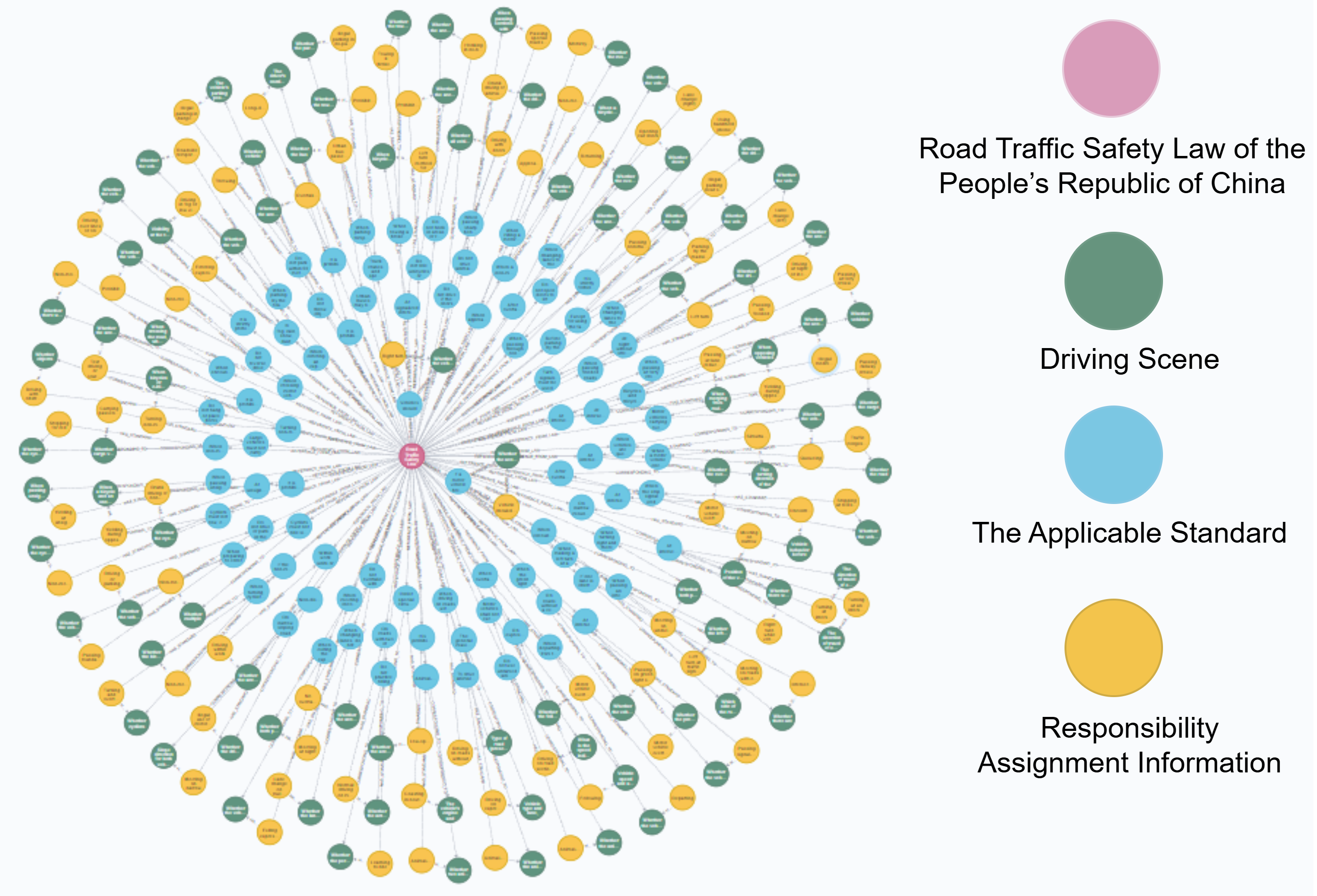}  
  \caption{Traffic Regulations Knowledge Graph in Neo4j}
  \label{fig:Fig11}
\end{figure}

Based on the Regulations for the Implementation of the Road Traffic Safety Law of the People's Republic of China \cite{gov}, we constructed the TRKG using the method described in Section \ref{sec31}, based on a large language model (LLM) \cite{openai_gpt4o_2024}. The knowledge graph includes three main types of nodes: 93 nodes representing Driving Scenes, 92 nodes capturing the corresponding Applicable Standards, and another 93 nodes detailing Responsibility Assignment Information. This structured representation ensures that each traffic scenario is comprehensively annotated with its relevant legal criteria. The TRKG is stored in a Neo4j graph database \cite{neo4j} for efficient querying and visualization, as illustrated in Fig. \ref{fig:Fig11}.
\\

\subsection{Accident Responsibility Classification Model Training}
To train an off-line responsibility assignment model, as described in Section \ref{sec32}, we first collected traffic accident image data. The ego vehicle was programmed to interact with the MetaDrive simulation environment, during which a series of accident scenarios were generated, capturing collisions between the ego vehicle and surrounding traffic. Each incident was then annotated with responsibility labels following the procedure outlined in Section \ref{sec32}, resulting in a dataset of image–label pairs, where each image corresponds to a specific liability category assigned to the ego vehicle.

The constructed data set was subsequently divided into training and test sets. To enhance the generalizability of the model, Gaussian noise was added to the training images as a data augmentation strategy. The distribution of samples in different categories of responsibility in the training, validation, and test sets is presented in Table \ref{tab:liability_distribution}.

\begin{figure}[h]
    \centering
    \begin{subfigure}{0.49\textwidth}
        \centering
        \includegraphics[width=\linewidth]{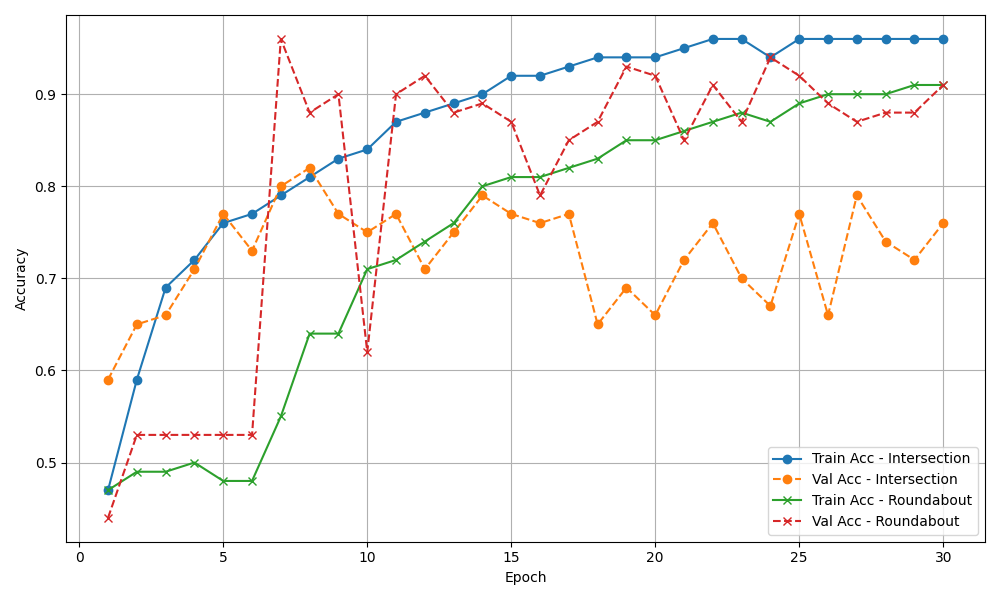}  
        \caption{Training and validation accuracy}
        \label{fig:aa}
    \end{subfigure} \hfill
    \begin{subfigure}{0.49\textwidth}
        \centering
        \includegraphics[width=\linewidth]{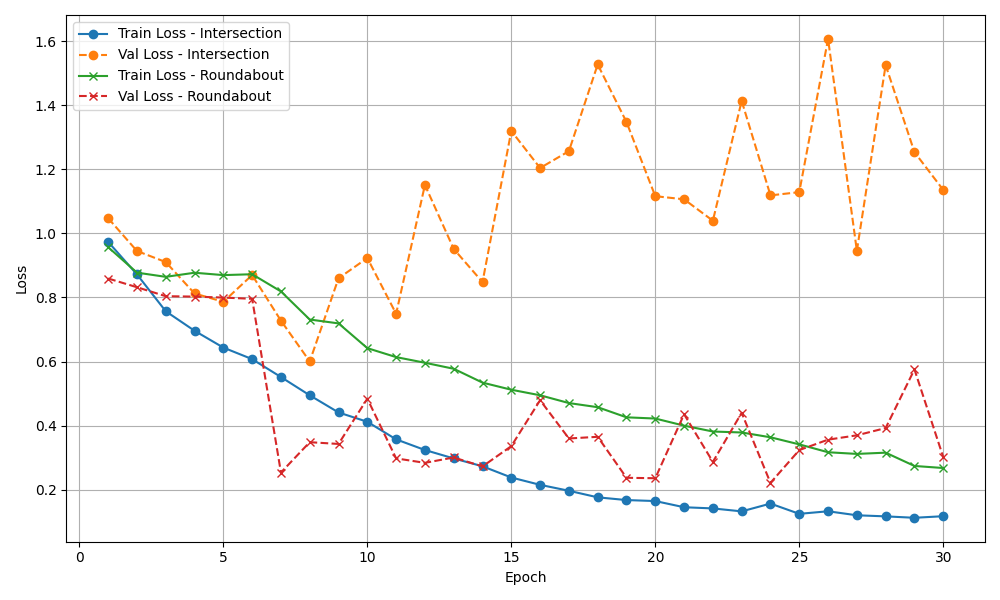}  
        \caption{Training and validation loss}
        \label{fig:bb}
    \end{subfigure}
    \caption{Training and validation accuracy and loss curves of the VIT model}
    \label{fig:vit_train}
\end{figure}

\begin{table}[h]
\centering
\caption{number of cases by liability category in training and test sets}
\label{tab:liability_distribution}
\begin{tabular}{l|c|c|c}
\textbf{Dataset} & \textbf{Primary Liability} & \textbf{Shared Liability} & \textbf{Secondary Liability} \\
\hline
Intersection Training Set & 3666 & 688 & 3342 \\ \hline
Intersection Validation Set  & 60   & 9   & 32   \\
\hline
Roundabout Training Set & 2982 & 288 & 2660 \\ \hline
Roundabout Validation Set  & 118   & 6   & 99   \\
\end{tabular}
\end{table}
We trained a Vision Transformer (ViT) model \cite{dosovitskiy2020image} to perform the classification task using the dataset described above. The corresponding training and validation loss and accuracy curves are depicted in Fig. \ref{fig:vit_train}. 

For the Intersection scenario, although the training accuracy consistently improved, the validation loss exhibited substantial fluctuations and even an increasing trend in later epochs. This instability may primarily result from the imbalanced and limited validation set. Specifically, the validation set included only 9 samples from the shared liability category and 32 from the secondary liability category, causing the model's generalization to be particularly sensitive to minor shifts in prediction distribution. Consequently, the sparse data distribution likely contributed to unstable validation metrics and potentially unreliable loss evaluations, particularly as the model increasingly overfitted to the dominant primary liability class.

In contrast, training outcomes for the Roundabout scenario were notably more stable. Validation accuracy showed a rapid and sustained increase after the 7th epoch, accompanied by consistently lower validation loss values. This suggests that the model exhibited substantially stronger generalization capabilities under this scenario.\\

\subsection{Autonomous Driving Model Training}
\begin{figure}[h]
  \centering
  \begin{minipage}[t]{0.48\textwidth}
    \centering
    \includegraphics[width=\textwidth]{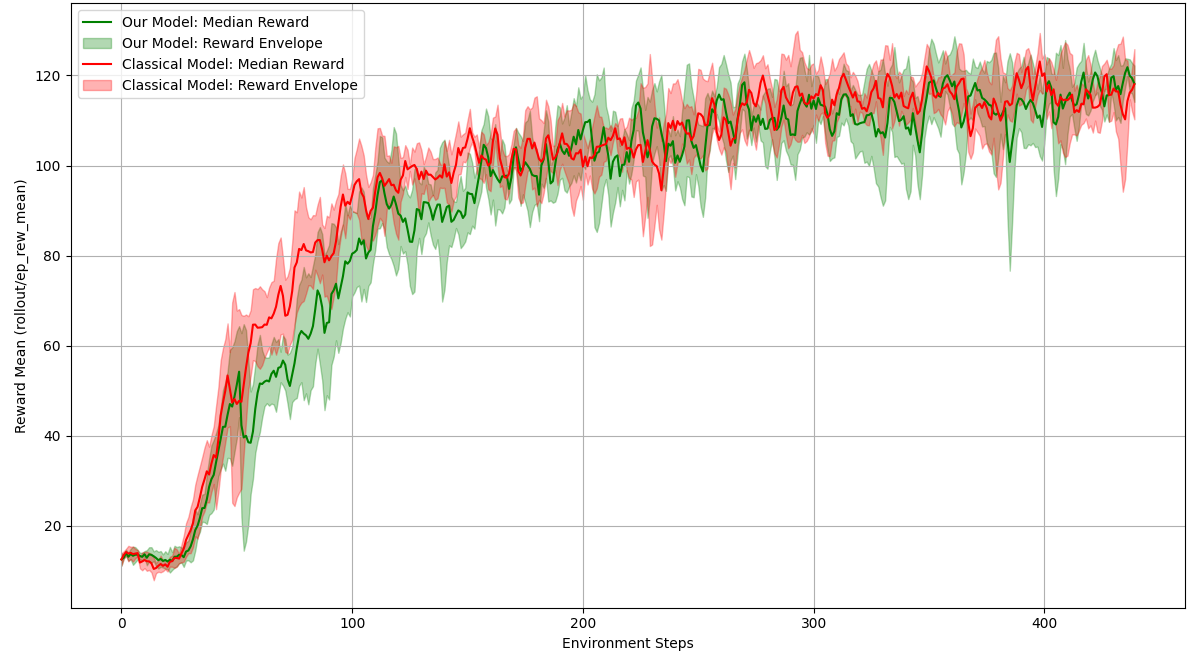}
    \caption*{(a) Convergence in the Scene of Intersection}
  \end{minipage}
  \hfill
  \begin{minipage}[t]{0.48\textwidth}
    \centering
    \includegraphics[width=\textwidth]{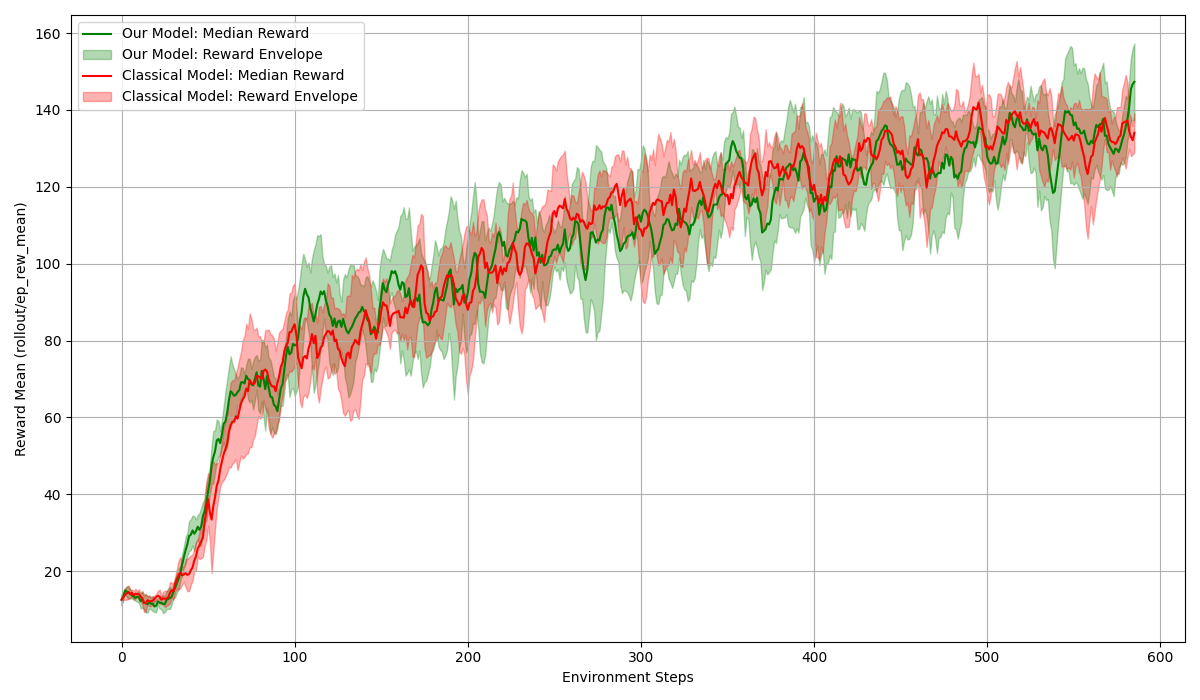}
    \caption*{(b) Convergence in the Scene of Roundabout}
  \end{minipage}
  \caption{Convergence Curves of PPO Training Across Scenarios}
  \label{fig:Fig4}
\end{figure}

We used the MetaDrive simulator \cite{li2022metadrive} to train the autonomous driving agent using the PPO algorithm in scenarios such as intersection and roundabout, based on the reward function described in Section \ref{sec34}. For both the baseline and our method, we conducted five repeated training runs in each scenario, resulting in a total of 20 experiments. The convergence of the model is shown in Fig. \ref{fig:Fig4}.\\

\subsection{Model Evaluation}
\subsubsection{Quantitative Analysis of the Automatic Driving Model}

We compare the autonomous driving model trained using the reward function proposed in this paper with a baseline model. The key difference between the two models lies in the reward function: the baseline model does not assign responsibility for traffic accidents, instead applying a uniform penalty of -20 for any agent involved in a collision. To ensure the robustness of the results, both our method and the baseline were each trained five times in both the intersection and roundabout scenarios, totaling 20 independent experiments. Table \ref{tab:comparison_main} summarizes the average outcomes using abbreviated metrics for clarity: CR (Crash), TO (Timeout), SU (Success), PR (Primary Responsibility), SH (Shared Responsibility), and SE (Secondary Responsibility). As shown, our model achieves higher success rates and bears significantly less primary responsibility in both scenarios compared to the baseline.

\begin{table}[h]
\centering
\caption{Comparison of Final State and Responsibility Distribution}
\label{tab:comparison_main}
\begin{tabular}{c|c|c|c|c|c|c|c}
\multirow{2}{*}{\textbf{Scenario}} & \multirow{2}{*}{\textbf{Method}} & \multicolumn{3}{c|}{\textbf{Final State}} & \multicolumn{3}{c}{\textbf{Responsibility Distribution}} \\
 & & \textbf{CR~(↓)} & \textbf{TO~(↓)} & \textbf{SU~(↑)} & \textbf{PR~(↓)} & \textbf{SH~(↓)} & \textbf{SE~(↑)} \\
\hline
\multirow{2}{*}{Intersection} 
 & Ours & \textbf{26.8\%} & \textbf{0\%} & \textbf{73.2\%} & \textbf{43.5\%} & \textbf{4.7\%} & \textbf{51.8\%} \\
 & Baseline & 35.0\% & \textbf{0\%} & 65.0\% & 57.0\% & 8.8\% & 34.2\% \\
\hline
\multirow{2}{*}{Roundabout} 
 & Ours & \textbf{45.6\%} & \textbf{0.4\%} & \textbf{54.0\%} & \textbf{50.8\%} & \textbf{11.3\%} & \textbf{37.9\%} \\
 & Baseline & 56.8\% & \textbf{0.4\%} & 42.8\% & 56.5\% & 13.9\% & 29.6\% \\
\end{tabular}
\end{table}

\subsubsection{Qualitative Analysis of the Automatic Driving Model}

We conducted a qualitative analysis of scenarios not included in the MetaDrive training set, comparing the performance of our model and the baseline model in these scenarios.

\begin{figure}[htbp]
    \centering
    \includegraphics[width=0.19\textwidth]{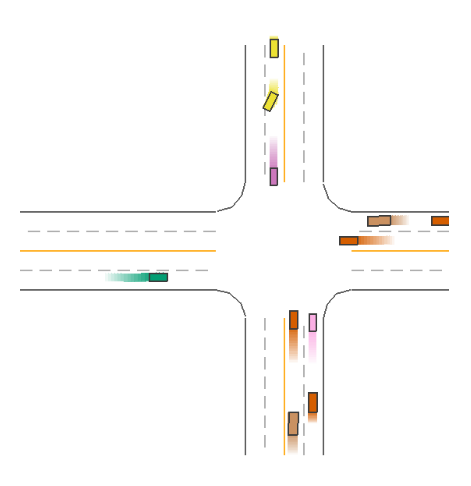}
    \includegraphics[width=0.19\textwidth]{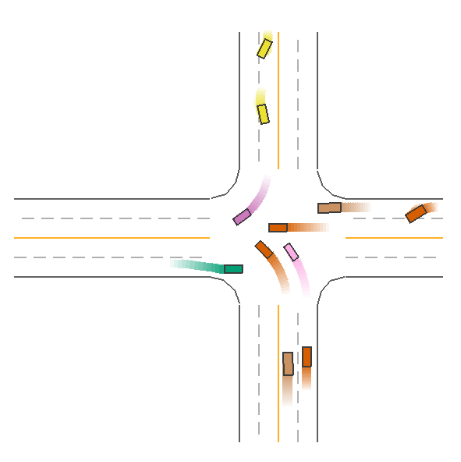}
    \includegraphics[width=0.19\textwidth]{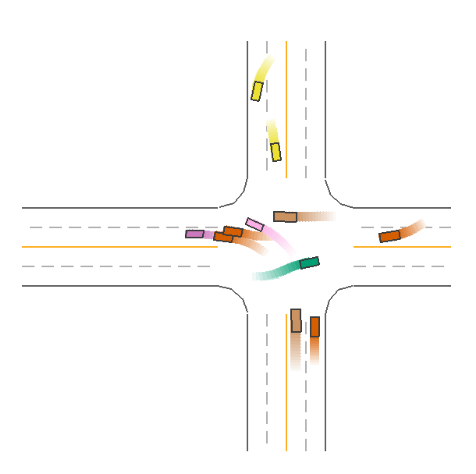}
    \includegraphics[width=0.19\textwidth]{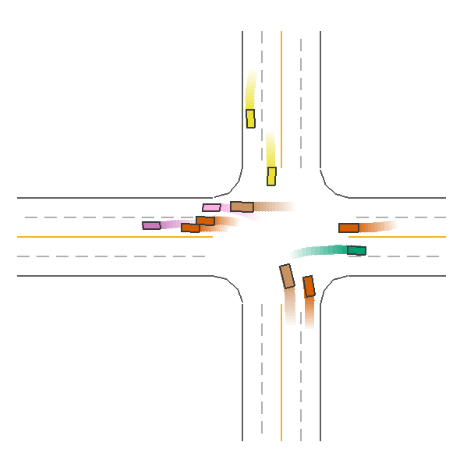}
    \includegraphics[width=0.19\textwidth]{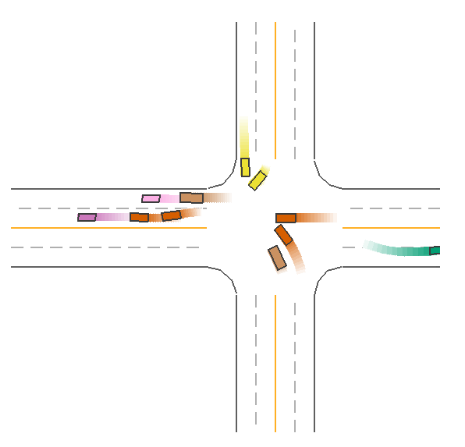}
    \caption{Selected key frames from the straight-driving scenario, showing decisive forward motion. The ego vehicle is colored green in each frame for clear identification.}
    \label{fig:straight_sequence}
\end{figure}

In the Crossroads scenario, our model tends to proceed straight through more decisively rather than hesitating. This behavior is attributed to the model's understanding of right-of-way rules: in such cases, traffic regulations generally require vehicles approaching from the right to yield. Therefore, if a collision occurs, the model anticipates that the oncoming vehicle would be primarily responsible, reinforcing its decision to proceed. As illustrated in Fig.~\ref{fig:straight_sequence}, the ego vehicle (green car) proceeds forward through the intersection with a steady trajectory.

\begin{figure}[htbp]
    \centering
    \includegraphics[width=0.19\textwidth]{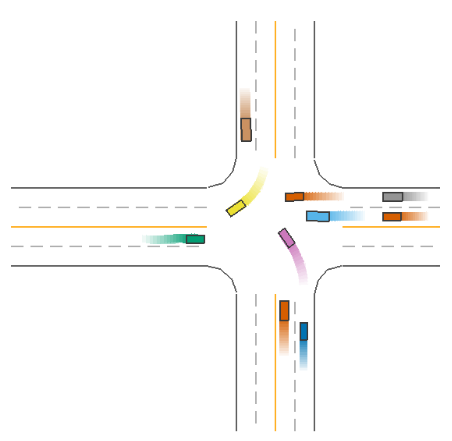}
    \includegraphics[width=0.19\textwidth]{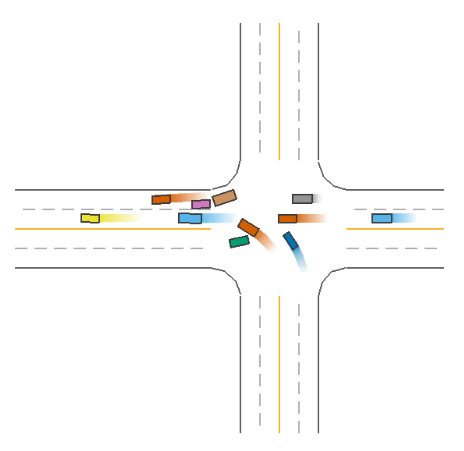}
    \includegraphics[width=0.19\textwidth]{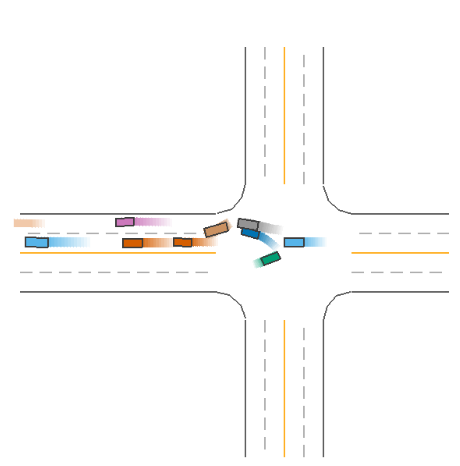}
    \includegraphics[width=0.19\textwidth]{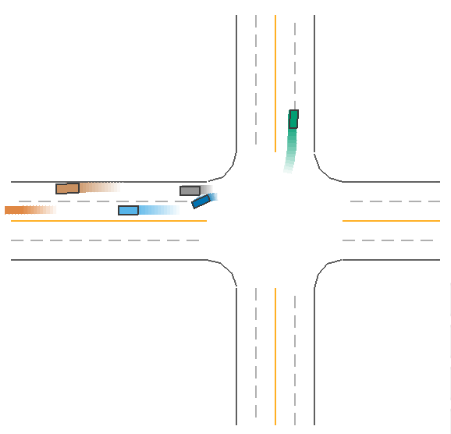}
    \includegraphics[width=0.19\textwidth]{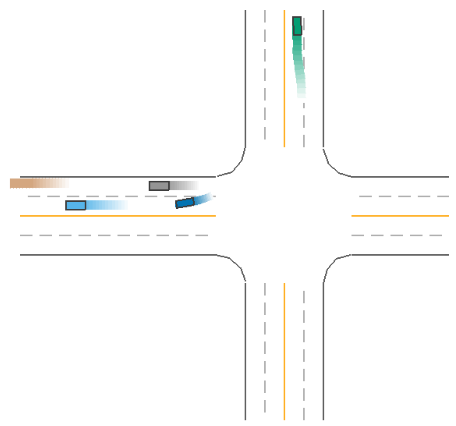}
    \caption{Selected key frames from the left-turn scenario, illustrating the vehicle’s cautious waiting behavior. The ego vehicle is colored green in each frame for clear identification.}
    \label{fig:left_turn_sequence}
\end{figure}

When performing the left-turn task, our model tends to wait for oncoming traffic to pass before proceeding, rather than risk proceeding early because the model believes it would bear full responsibility if a collision occurs with oncoming straight traffic. This conservative behavior is visualized in the sequence shown in Fig.~\ref{fig:left_turn_sequence}, which captures the temporal progression of the left-turn maneuver.

\begin{figure}[htbp]
    \centering
    \includegraphics[width=0.19\textwidth]{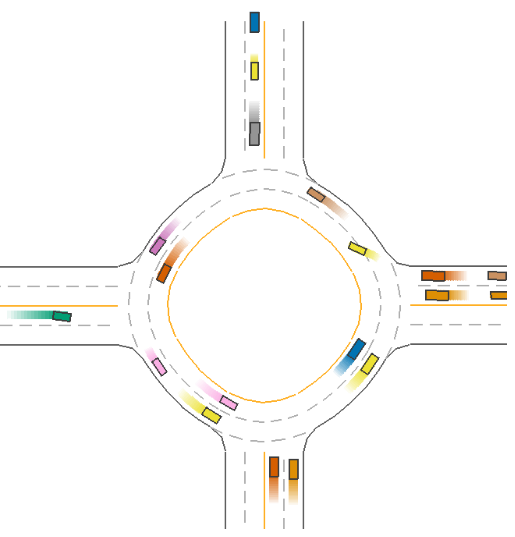}
    \includegraphics[width=0.19\textwidth]{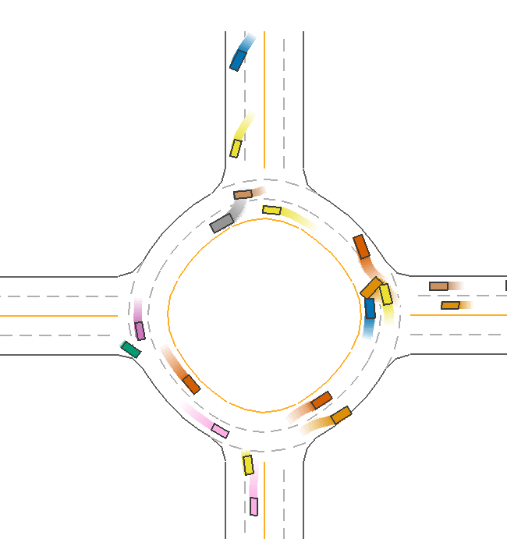}
    \includegraphics[width=0.19\textwidth]{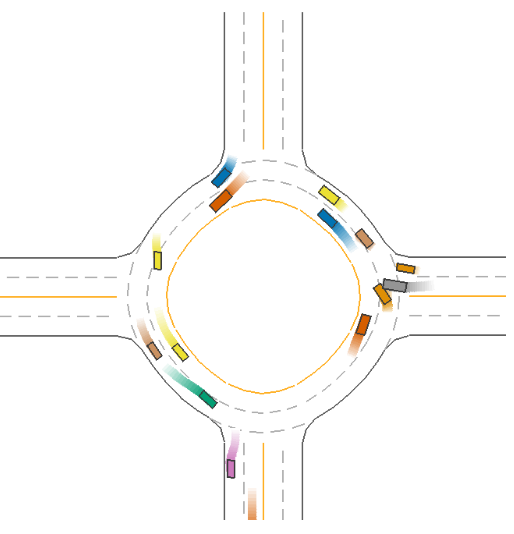}
    \includegraphics[width=0.19\textwidth]{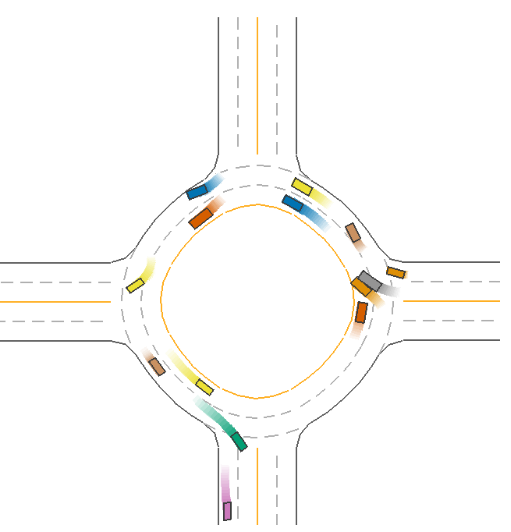}
    \includegraphics[width=0.19\textwidth]{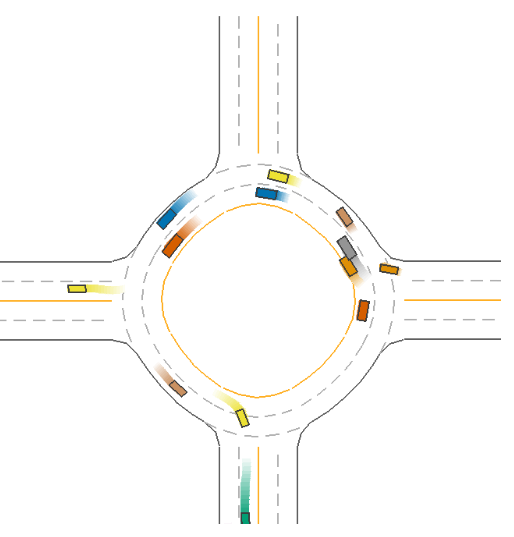}
    \caption{Selected key frames from the roundabout scenario, demonstrating yield-before-entry. The ego vehicle is colored green in each frame for clear identification.}
    \label{fig:roundabout_sequence}
\end{figure}

In the roundabout scenario, our model exhibits a distinct behavioral pattern: It yields to vehicles already circulating within the roundabout before entering. We attribute this behavior to the model’s adherence to the traffic regulation that stipulates that vehicles entering the roundabout must yield to those already circulating within it. Fig.~\ref{fig:roundabout_sequence} presents a visual sequence of this behavior during a representative roundabout traversal.

\subsubsection{Ablation Study}
We compared RAG-based responsibility assignment methods, as proposed in Section \ref{sec32}, and the baseline. For the baseline method, we use GPT-4o from Open AI \cite{openai_gpt4o_2024}. Through experimentation, we obtained the accuracy, precision, recall and F1 score of these two methods, as shown in Table \ref{tab:comparison}. The results indicate that the proposed RAG-based method outperforms the baseline method in terms of responsibility assignment accuracy for road traffic accident tasks.

\begin{table}[ht]
\caption{Comparison of Responsibility Assignment Performance}
\centering
\label{tab:comparison}
\begin{tabular}{c|c|c|c}
\textbf{Metric} & \textbf{Method} & \textbf{Intersection} & \textbf{Roundabout} \\
\hline
\multirow{2}{*}{Accuracy} 
 & Ours & \textbf{79.21\%} & \textbf{78.03\%} \\
 & GPT-4o & 63.37\% & 68.61\% \\
\hline
\multirow{2}{*}{Precision} 
 & Ours & \textbf{79.28\%} & \textbf{79.49\%} \\
 & GPT-4o & 68.28\% & 69.99\% \\
\hline
\multirow{2}{*}{Recall} 
 & Ours & \textbf{79.21\%} & \textbf{78.03\%} \\
 & GPT-4o & 63.37\% & 68.61\% \\
\hline
\multirow{2}{*}{F1 Score} 
 & Ours & \textbf{79.20\%} & \textbf{78.58\%} \\
 & GPT-4o & 64.36\% & 68.94\% \\
\end{tabular}
\end{table}

\section{Conclusion}
This study proposes a novel paradigm for the design of reward functions for autonomous driving, grounded in formal traffic regulations. By constructing a comprehensive and structured traffic regulation knowledge graph, the framework enables reinforcement learning agents to effectively internalize legal norms. The approach is further enhanced through the integration of VLMs and RAG, facilitating dynamic adaptation and real-time updating of reward functions via rule retrieval. Experimental results in various simulated driving scenarios demonstrate that the proposed method improves regulatory compliance while reducing the incidence of hazardous behaviors.\\
Future research may focus on incorporating more granular and region-specific traffic laws to improve the generalizability of the model in heterogeneous legal systems.

\section*{Acknowledgments}
This work was supported in part by the National Natural Science Foundation of China under Grant No. 52402504, in part by the Aero Engine Corporation of China Industry-University-Research Collaboration Project under Grant No. HFZL2024CXY003, and in part by the Aeronautical Science Foundation of China under Grant No. 2024L039057003.


\bibliography{references}  

\end{document}